\documentclass[letterpaper, 10 pt, conference]{ieeeconf}  % Comment this line out if you need a4paper

\IEEEoverridecommandlockouts                              % This command is only needed if 
\usepackage{graphicx}
\usepackage{algorithmic}
\usepackage{amsmath} % assumes amsmath package installed
\usepackage{amssymb}  % assumes amsmath package installed
\usepackage{subfigure}
\usepackage{float}

\title{\LARGE \bf
Does Deep Learning REALLY Outperform Non-deep Machine Learning for Clinical Prediction on Physiological Time Series?
}

\author{Ke Liao$^{1,2}$, Wei Wang$^{1}$, Armagan Elibol$^{2}$, Lingzhong Meng$^{3}$, Xu Zhao$^{4}$, and Nak Young Chong$^{2}$% <-this % stops a space
\thanks{$^{1}$K. Liao and W. Wang are with the Ricoh Software Research Center (Beijing) Co., Ltd., Beijing, China ({\tt\small \{ke.liao, wei.wang5\}@cn.ricoh.com}).}%
\thanks{$^{2}$K. Liao is a Ph.D. student of the Japan Advanced Institute of Science and Technology (JAIST). A. Elibol and N.Y. Chong are with the School of Information Science, JAIST, Ishikawa, Japan ({\tt\small \{s1820033, aelibol, nakyoung\}@jaist.ac.jp}).}%
\thanks{$^{3}$L. Meng is with the Department of Anesthesiology, Yale University School of Medicine, New Haven, CT 06520, USA ({\tt\small lingzhong.meng@yale.edu}).}%        
\thanks{$^{4}$X. Zhao is with the Department of Anesthesiology, The Second Xiangya Hospital, Central South University, Changsha, Hunan Province, China ({\tt\small zhaoxu109@hotmail.com}).}%   
}

\begin{document}

\maketitle
\thispagestyle{empty}
\pagestyle{empty}

%%%%%%%%%%%%%%%%%%%%%%%%%%%%%%%%%%%%%%%%%%%%%%%%%%%%%%%%%%%%%%%%%%%%%%%%%%%%%%%%
\begin{abstract}

Machine learning has been widely used in healthcare applications to approximate complex models, for clinical diagnosis, prognosis, and treatment. As deep learning has the outstanding ability to extract information from time series, its true capabilities on sparse, irregularly sampled, multivariate, and imbalanced physiological data are not yet fully explored. In this paper, we systematically examine the performance of machine learning models for the clinical prediction task based on the EHR, especially physiological time series. We choose Physionet 2019 challenge public dataset \cite{reyna2019early} to predict Sepsis outcomes in ICU units. Ten baseline machine learning models are compared, including 3 deep learning methods and 7 non-deep learning methods, commonly used in the clinical prediction domain. Nine evaluation metrics with specific clinical implications are used to assess the performance of models. Besides, we sub-sample training dataset sizes and use learning curve fit to investigate the impact of the training dataset size on the performance of the machine learning models. We also propose the general pre-processing method for the physiology time-series data and use Dice Loss \cite{vicar2019sepsis} to deal with the dataset imbalanced problem. The results show that deep learning indeed outperforms non-deep learning, but with certain conditions: firstly, evaluating with some particular evaluation metrics (AUROC, AUPRC, Sensitivity, and FNR), but not others; secondly, the training dataset size is large enough (with an estimation of a magnitude of thousands).

\end{abstract}

%%%%%%%%%%%%%%%%%%%%%%%%%%%%%%%%%%%%%%%%%%%%%%%%%%%%%%%%%%%%%%%%%%%%%%%%%%%%%%%%
\section{INTRODUCTION}

With the widespread adoption of Electronic Health Records (EHR), there is an increased emphasis on predictive models that can help clinical decisions and treatment recommendation. EHR contains various types of data, including demographics information, laboratory values, vital signs, physiology signals and treatment records, which are generally sparse, irregularly sampled, and multivariate. With the development of machine learning methods, especially deep learning and neural networks, more research focuses on clinical prediction or physiology management estimation with EHR using machine learning methods \cite{harutyunyan2019multitask}. Deep learning shows excellent capabilities to extract information and features from big data, but the performance of deep learning depends on the training dataset size, model structure, data quality, and similar other related conditions. Many research or actual work fails to reach the ideal target when utilizing deep learning models \cite{RN4}. In some cases, non-deep learning, such as XGBoost has a similar or even better performance with deep neural networks \cite{wang2019prediction}. 

In this paper, we systematically analyze the machine learning models' performance on the clinical prediction domain, using EHR, especially the physiology time series data. We propose a general framework to handle the raw multivariate time series data. We utilize the Physionet 2019 challenge public dataset \cite{reyna2019early} to predict Sepsis outcomes in ICU units, defined as a binary classification problem. And we train and evaluate the datasets with 10 baseline machine learning models: 3 deep learning models: (1) InceptionTime \cite{fawaz2020inceptiontime}, (2) TCN (Temporal CNN) \cite{kok2020automated} and (3) Transformer \cite{wu2020deep} and 7 non-deep learning models: (4) LR (Logistic Regression), (5) SVM (Support Vector Machines), (6) KNN (K Nearest Neighbors), (7) GP (Gaussian Process), (8) DT (Decision Tree), (9) RF (Random Forests) and (10) XGBoost \cite{wang2019prediction}. Then, we evaluate the performance of different machine learning models using 9 different evaluation metrics with specialized clinical meanings. Also, as the training dataset size is a crucial aspect for the models' performance, we sub-sample the training dataset and present a quantitative analysis of the relationship between the training dataset size and the performance of machine learning models using the learning curve fit method.

Based on the proposed systematic and quantitative analysis framework, we find that deep learning \textbf{indeed} outperforms non-deep learning only under certain conditions. 
\begin{itemize}
\item The four evaluation metrics: AUROC, AUPRC, Sensitivity, and FNR show that deep learning outperforms non-deep learning. For example, AUROC is greater than 0.9 for deep learning models, while it is less than 0.8 for non-deep learning models. 
\item Labeled training dataset size: The larger training dataset improves the performance of the machine learning models, and deep learning models need a certain number of training dataset sizes to reach superior performance. The threshold value depends on different evaluation metrics. The overall estimate of training dataset sizes is about a magnitude of thousands. 
\end{itemize}

\section{RELATED WORK}
There is an extensive body of clinical prediction based on EHR data, especially the physiology time series. Hatib \emph{et al.}~\cite{hatib2018machine} use feature engineering to extract features from arterial pressure waveform to the prediction of an upcoming hypotensive event. Lee \emph{et al.}~\cite{lee2018development} extract basic statistical features and then uses feedforward neural networks to predict mortality.  

Some recent works try to extract the information or features from raw time series data, without feature engineering or statistical analysis. Kok \emph{et al.}~\cite{kok2020automated} use TCN for automated prediction of Sepsis outcome. Che \emph {et al.}~\cite{vicar2019sepsis} develop LSTM for Sepsis prediction with missing values in raw data. Fawaz \emph{et al.}~\cite{fawaz2020inceptiontime} provide InceptionTime (a CNN-based model) to deal with time series classification problem. Zhao \emph{et al.}~\cite{RN4} use the InceptionTime model to analyze the intraoperative time-series monitoring data, predicting the post-hysterectomy quality of recovery. Wu \emph{et al.}~\cite{wu2020deep} focus on the Transformer model and use it in the influenza prevalence prediction case.

\section{METHODS}

\subsection{Data analysis and preprocessing}

\subsubsection{Data characteristics}
This work is motivated by the analysis of physiological time series data in EHR, which are generally sparse, irregularly sampled, and multivariate. The EHR data contains the real-time physiological status of patients~\cite{harutyunyan2019multitask}, such as vital signs or other physiology signals, laboratory values, demographics information, treatments, and outcomes. The data contains continuous and categorical data with varied time lengths and different sampling frequencies. The physiology time series are usually sparse, with missing data both in the time dimension and feature dimension. Also, the positive sample size is generally much smaller than the negative one in most clinical prediction domains. The imbalanced dataset is a fundamental problem in the clinical prediction field. 

\subsubsection{Public datasets}

Physionet 2019 challenge dataset (early prediction of Sepsis from clinical data) includes hourly physiological time series data consisting of 8 vital signs, 26 laboratory values, and 6 demographic details of 40,336 samples from the MIMIC III database \cite{reyna2019early}. The time length varies from 8 to 336 time-steps. The outcome is the onset of sepsis. %as given in Fig.~\ref{fig:datasetA}.
%
%\begin{figure}[!ht]%[!ht]
%\begin {center}
%\includegraphics[width=0.5\textwidth]{img/datasetA.png}
%\caption{One patient’s raw data from the Physionet 2019 %challenge dataset.}
%\label{fig:datasetA}
%\end {center}
%\end{figure}

\subsubsection{Preprocessing}

\paragraph{Missing data}

The physiology data is inconsistent in the timeline. Meanwhile, missing values are correlated with feature labels. As the physiology data reflect the clinical status of patients, we assume that the values vary little within a small time window. Thus, we use forward and backward fill for the missing points in the time domain. When the missing value relates to the feature labels, we fill other specific values with a special meaning, \emph{e.g.}, -1 or N/A.
%Fig.~\ref{fig:missing} visualizes the data of one sample from the Sepsis prediction dataset. The black block refers to the missing data. 

%\begin{figure}[!ht]%[!ht]
%\begin {center}
%\includegraphics[width=0.5\textwidth]{img/missing_sepsis.PNG}
%\caption{Values distribution of all the features.}
%\label{fig:missing}
%\end {center}
%\end{figure}

\paragraph{Various data length}

Data length depends on time length and sampling frequencies. The time lengths of physiology time series are significantly different among features or patients. Also, the sampling frequencies are different among physiology features. The data filling process includes two steps:  1) firstly, utilize forward and backward filling according to the max sampling frequency. 2) fill each sample according to the max length of all samples with a special meaning, \emph{e.g.}, -1 or N/A.

\paragraph{Continuous and categorical data}

Generally, vital signs and laboratory values are continuous data and we use normalization to maintain the contribution of each feature and keep the model unbiased. Demographics information is mainly categorical data and we use one-hot encoding to convert the categorical data. After the filling, normalization, and one-hot encoding, processed EHR data is formatted for training and testing.

\paragraph{Imbalanced dataset}

The imbalance issue is a common problem in the clinical domain: positive samples are much less than negative ones. For example, the Sepsis predction dataset has 40,336 samples with 2,932 positive patients, with a rate of 7.27\%. Two main methods deal with the imbalance dataset: data augmentation and special loss functions. As the data augmentation methods affect the dataset size, we use Dice Loss~\cite{vicar2019sepsis} as the loss function of the prediction problem.

\subsection{Models}

\subsubsection{Deep learning models}
Deep learning models are based on neural networks that are composed of a large number of layers and parameters, which can be potentially used for extracting features from time series and downstream tasks such as regression, classification, forecasting, and feature embedding. We employ 3 different deep learning models to directly tackle the physiology time series for Sepsis prediction task, including InceptionTime \cite{fawaz2020inceptiontime}, TCN (Temporal CNN) \cite{kok2020automated} and Transformer \cite{wu2020deep}.

\paragraph{InceptionTime}
InceptionTime is a 1D-CNN-based deep neural network. It consists of a series of Inception modules followed by a Global Average Pooling layer and a Dense layer with a softmax activation function. %, as shown in Fig.~\ref{fig:InceptionTime}. 
In each Inception module, multi-dimension time series data are transformed into 1D data. Three one-dimension filters with lengths of 10, 20, and 40, respectively, are applied simultaneously to the output. Residual blocks are added to mitigate the vanishing gradient problem.

%\begin{figure}[!ht]%[!ht]
%\begin {center}
%\includegraphics[width=0.45\textwidth]{img/inceptionTime.PNG}
%\caption{Architecture of the InceptionTime model. \cite{RN4}}
%\label{fig:InceptionTime}
%\end {center}
%\end{figure}

\paragraph{TCN}
TCN is a variant of the CNN and employs ordinary convolutions and dilations. It is suitable for sequential data with temporality and large receptive fields. The latent correlation among series can be learned from this model. %Fig.~\ref{fig:TCN} shows the basic structure of TCN for time series data.

%\begin{figure}[!ht]%[!ht]
%\begin {center}
%\includegraphics[width=0.45\textwidth]{img/TCN.PNG}
%\caption{Architecture of the TCN model. %\cite{kok2020automated}}
%\label{fig:TCN}
%\end {center}
%\end{figure}

\paragraph{Transformer}

The Transformer-based time series model consists of Transformer encoder layers.%, shown in Fig.~\ref{fig:Transformer}.
The network is composed of an input layer, a positional encoding layer, and a stack of transformer encoder layers. We use one transformer encoder layer in this work.

%\begin{figure}[!ht]%[!ht]
%\begin {center}
%\includegraphics[width=0.45\textwidth]{img/Transformer.PNG}
%\caption{Architecture of the Transformer model. %\cite{zerveas2020transformer}}
%\label{fig:Transformer}
%\end {center}
%\end{figure}

\subsubsection{Non-deep learning models}
For Non-deep learning models, we evaluate 7 models with the statistical information extracted from raw time series data. Some machine learning models are inconvenient to be used with raw time-series data, such as logistic regression. Usually, the preprocessing is needed to extract the basic (descriptive) statistical information or feature from the time-series data to be used in the machine learning methods~\cite{hatib2018machine},\cite{lee2018development}. The statistical information includes the maximum, minimum, average, median, standard deviation, quantiles, moments, \emph{etc}.

\subsection{Training and Evaluation}

\subsubsection{Training process}
We choose 36 features (8 vital signs, 26 laboratory values, and 2 demographic details) to predict the sepsis outcome as a binary classification. We divided the training dataset into training sets and test sets. The test sets have 4,033 samples (10\% of all dataset sizes) and the training sets have 36,303 samples (90\% of all dataset sizes). Also, the training sets are sub-sampled from 100 to 36,303 sample size: 100, 200, 400, 600, 800, 1000, 2000, 4000, 6000, 8000, 10000, 15000, 20000, 25000, 30000, 36303.

\subsubsection{Performance metrics and evaluation}
For the classification task, the most commonly reported metric is the area under the receiver operator characteristic curve (AUROC), which combines the information from sensitivity and specificity into a single value. We also report AUPRC since it is a good metric for describing model performance given imbalanced datasets. And we evaluate the accuracy, sensitivity (recall, TPR (true positive rate)), specificity (TNR, 1-FPR (false positive rate)), FNR, FPR, NPV, PPV, which represent actual clinical significance \cite{wong2011measures}.

\subsubsection{Training dataset size impact}

The evaluation results of machine learning models are largely dependent on the training dataset size. We choose the learning curve approach~\cite{cho2015much} to explore the relationship between the performance and the training dataset size. The curve model is represented as an inverse power law function given by

\begin{equation}
y=f(x,\boldsymbol{b})=100+b_{1}\cdot x^{b_{2}}.
\label{eq_lc}
\end{equation}

Using the evaluation results of the models trained with different training datasets sizes, we calculate the unknown parameters (b1 and b2) with nonlinear weighted least squares optimization using the following equation.

\begin{equation}
\begin{aligned}
E(\boldsymbol{b})
&=\sum_{p=1}^{m}(\omega _{p}\cdot (t_{p}-y_{p}))^{2} \\
&=\sum_{p=1}^{m}(\omega _{p}\cdot (t_{p}-f(x_{p},\boldsymbol{b})))^{2} \\
&=\sum_{p=1}^{m}(\omega _{p}\cdot r_{p}(\boldsymbol{b})^{2}) \\
&=\boldsymbol{R}^{T}W\boldsymbol{R},
\end{aligned}
\label{eq_optim}
\end{equation}
where $r_p$ is the residual between the real value and the fitted function. 

\section{RESULT AND DISCUSSION}

\subsection{Comparison of machine learning models' performance}

Fig. 1 contains the evaluation results of the Sepsis prediction task (a binary classification problem) based on different machine learning models.

\begin{figure*}
  \begin{minipage}[b]{0.4\textwidth} 
  \parbox[][4cm][c]{\linewidth}{
    \centering 
    \includegraphics[width=0.9\textwidth]{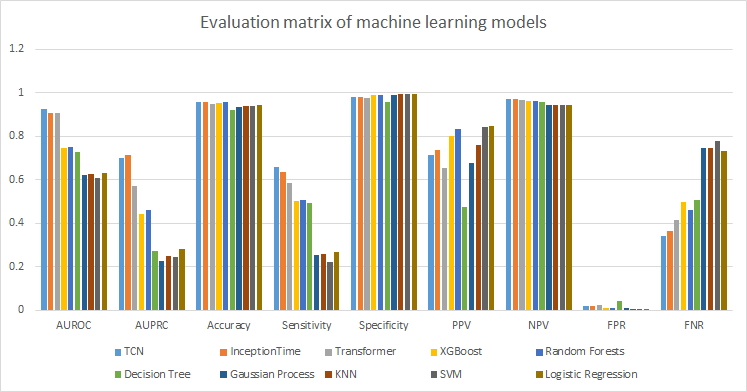} 
    \label{fig:comparison_result_A}
  }
  \end{minipage}% 
  \begin{minipage}[b]{0.6\textwidth} 
    \centering
    \scalebox{0.6}{
    \begin{tabular}{|l||c|c|c|c|c|c|c|c|l|}
\hline
\multicolumn{1}{|c||}{\textbf{Model}} & \textbf{AUROC} & \textbf{AUPRC} & \textbf{Accuracy} & \textbf{Sensitivity} & \textbf{Specificity} & \textbf{PPV} & \textbf{NPV} & \textbf{FPR} & \textbf{FNR} \\ \hline
\textbf{TCN} & \textbf{0.924}          & \textbf{0.698}         & 0.956             & \textbf{0.659}                & 0.979                & 0.715        & 0.973        & 0.021        & \textbf{0.341}        \\ \hline
\textbf{InceptionTime} & \textbf{0.907}          & \textbf{0.713}          & 0.957             & \textbf{0.635}                & 0.982                & 0.735        & 0.972        & 0.018        & \textbf{0.365}        \\ \hline
\textbf{Transformer}   & \textbf{0.909}          & \textbf{0.572}          & 0.947             & \textbf{0.584}                & 0.976                & 0.653        & 0.968        & 0.024        & \textbf{0.416}        \\ \hline
\textbf{XGBoost}       & 0.746          & 0.444          & 0.955             & 0.502                & 0.99                 & 0.803        & 0.962        & 0.01         & 0.498        \\ \hline
\textbf{Random Forests}                                   & 0.75           & 0.459          & 0.957             & 0.509                & 0.992                & 0.832        & 0.963        & 0.008        & 0.461        \\ \hline
\textbf{Decision Tree} & 0.726          & 0.271          & 0.923             & 0.495                & 0.957                & 0.474        & 0.96         & 0.043        & 0.505        \\ \hline
\textbf{Gaussian Process}                                 & 0.622          & 0.226          & 0.937             & 0.253                & 0.991                & 0.679        & 0.944        & 0.009        & 0.747        \\ \hline
\textbf{KNN} & 0.625          & 0.248          & 0.94              & 0.256                & 0.994                & 0.758        & 0.945        & 0.006        & 0.744        \\ \hline
\textbf{SVM} & 0.609          & 0.244          & 0.94              & 0.222                & 0.997                & 0.844        & 0.942        & 0.003        & 0.778        \\ \hline
\textbf{Logistic Regression}                              & 0.631          & 0.279          & 0.943             & 0.266                & 0.996                & 0.848        & 0.945        & 0.004        & 0.734        \\ \hline
\end{tabular}}
  \label{table:comparison_result_B}
  \end{minipage} 
  \label{fig:comparison_result}
  \caption{Performance comparison of machine learning models.}
\end{figure*}

Our findings include:
\begin{itemize}
\item The deep learning models outperform the non-deep learning methods for the AUROC, AUPRC, sensitivity, and FNR.
\item For the accuracy, specificity, NPV, and FPR, both deep learning models and non-deep learning models have similar performance results. 
\item The performance of PPV evaluation metric differs from diverse machine learning models.
\item All machine learning models could be divided into three groups based on performance: (1) deep learning models; (2) XGBoost, Random Forest, and Decision Tree; (3) Gaussian process, KNN, SVM, and logistic regression. The performance is ranked in the order of best to worst: Group 1, Group 2, Group 3.
\item Among non-deep learning models, XGBoost performs best compared to the other non-deep learning models. XGBoost could not perform as well as deep learning methods though.
\end{itemize}

\subsection{Training dataset size and learning curves}

\begin{figure*}
  \centering 
  \subfigure[AUROC]{ 
    \label{fig:dataset_auroc}
    \includegraphics[width=0.45\textwidth]{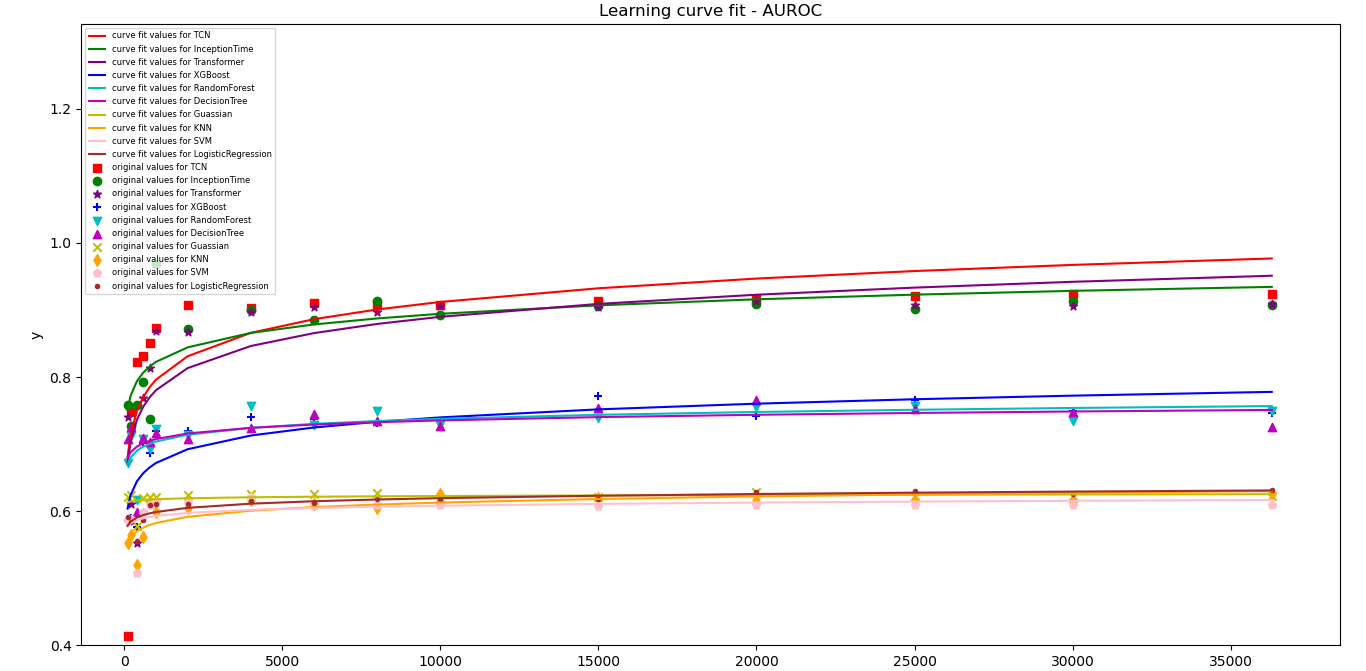} 
  } 
  \subfigure[AUPRC]{ 
    \label{fig:dataset_auprc} 
    \includegraphics[width=0.45\textwidth]{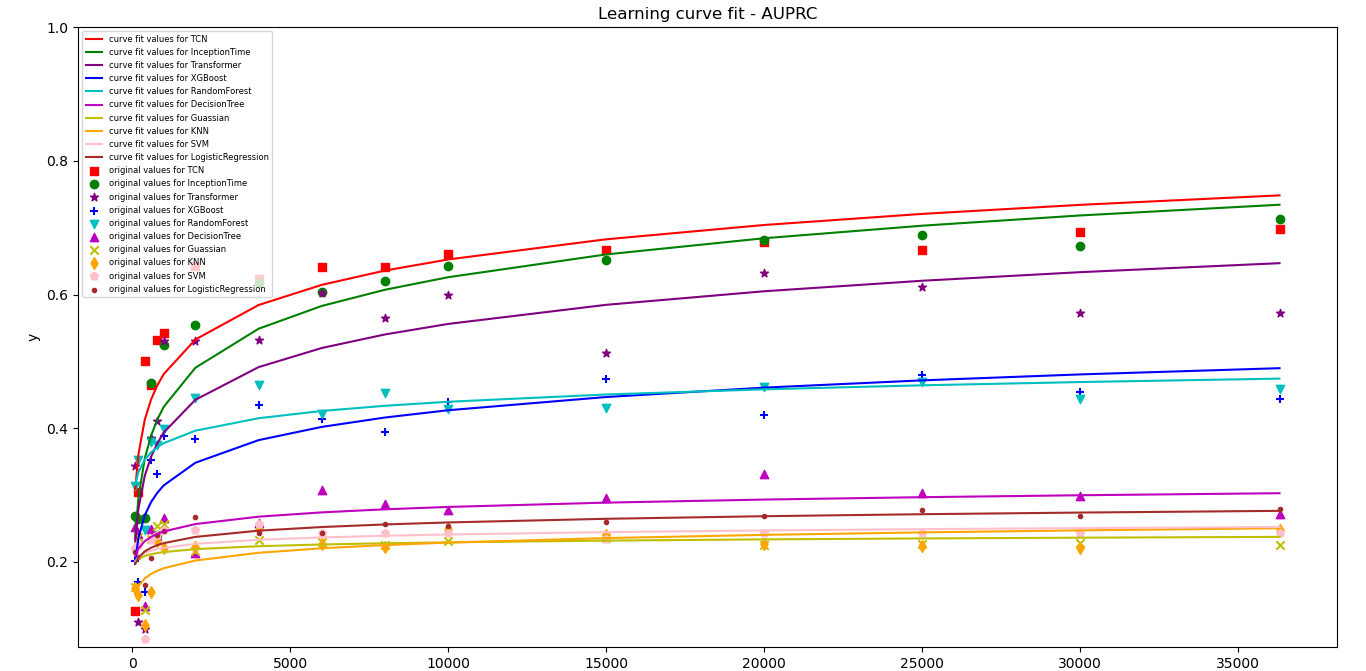} 
  }
  
  \subfigure[Sensitivity]{ 
    \label{fig:dataset_sensitivity}
    \includegraphics[width=0.45\textwidth]{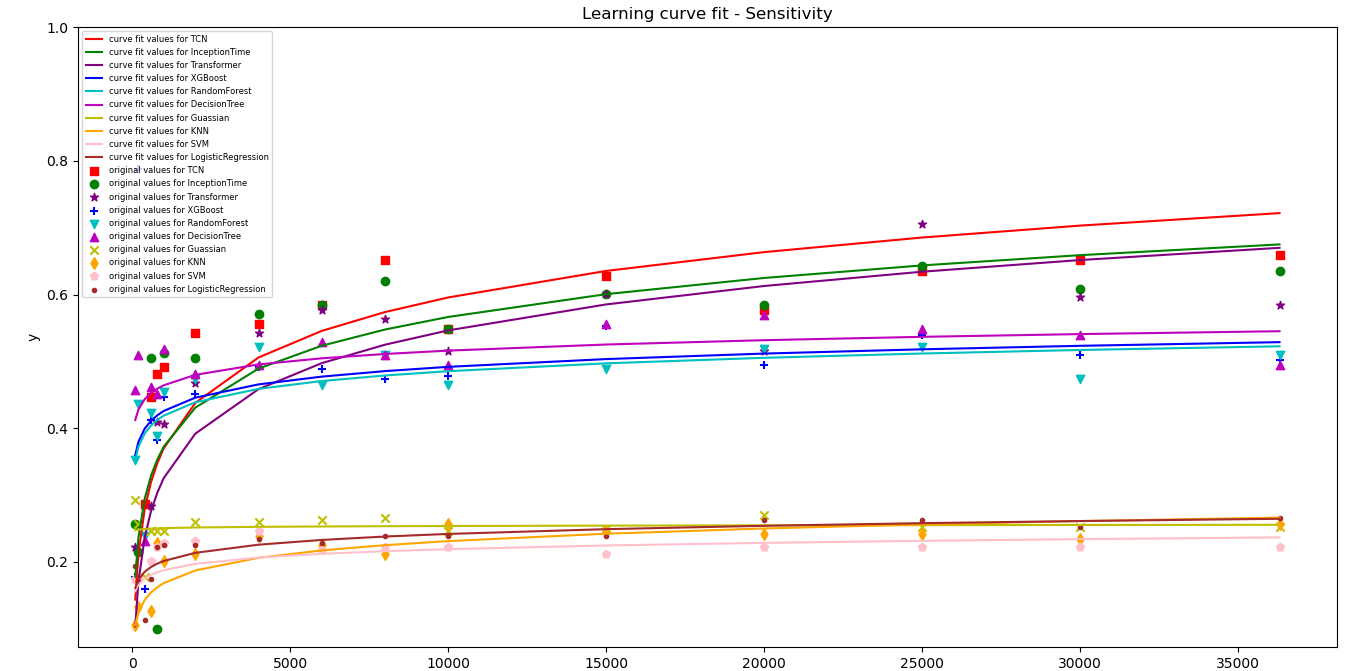} 
  } 
  \subfigure[FNR]{ 
    \label{fig:dataset_fnr} 
    \includegraphics[width=0.45\textwidth]{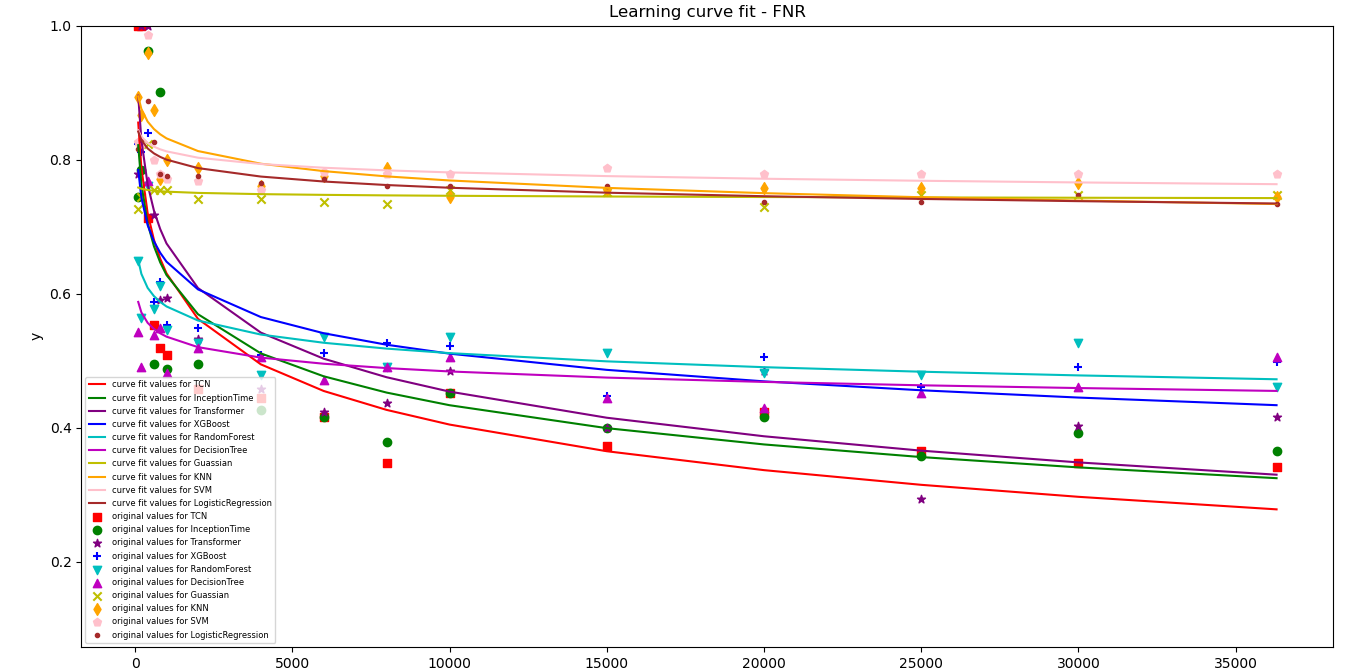} 
  } 
  \caption{AUROC, AUPRC, Sensivitity and FNR Learning Curve of different models.} 
  \label{fig:dataset_lc}
\end{figure*}

We also analyze the relationship between the training dataset size and the model performance. Fig.~\ref{fig:dataset_lc} depicts the impact of the training dataset size on the performance of the models, for the AUROC (upper left), AUPRC (upper right), Sensitivity (below left), and FNR (below right). Each line represents a model's performance according to the increasing number of training dataset sizes.

Our findings include:
\begin{itemize}
\item When the training dataset size is small, the deep learning models do not outperform the non-deep learning models, especially for the sensitivity evaluation metrics.
\item With the increased training dataset size, the deep learning models show an excellent ability comparing with non-deep learning models for all four evaluation metrics.
\item The increased number of training dataset size has a larger impact on Group 1 than Group 2, while no influence on Group 3.
\item With the increased training dataset size, all the models' performance approach saturation.
\end{itemize}

\section{CONCLUSIONS}

Our systematical and quantitative study showed that deep learning indeed outperforms non-deep learning with specific training dataset size (an estimation of a magnitude of thousands) and some particular evaluation metrics (AUROC, AUPRC, Sensitivity, and FNR). No general effect of machine learning models or deep learning models was observed with small sample sizes (below estimation of a magnitude of thousands). Therefore, deep learning has the potential to be used as a tool for clinical prediction.

The future direction of our work will extend our quantitative analysis to investigate the relationship between the clinical research performance and the scale of the network, data points, the sampling frequency of raw data, augmentation of data, \emph{etc}. It can assist researchers to determine the models and the dataset quality for clinical prediction research based on machine learning approaches.


%%%%%%%%%%%%%%%%%%%%%%%%%%%%%%%%%%%%%%%%%%%%%%%%%%%%%%%%%%%%%%%%%%%%%%%%%%%%%%%%

%%%%%%%%%%%%%%%%%%%%%%%%%%%%%%%%%%%%%%%%%%%%%%%%%%%%%%%%%%%%%%%%%%%%%%%%%%%%%%%%

%%%%%%%%%%%%%%%%%%%%%%%%%%%%%%%%%%%%%%%%%%%%%%%%%%%%%%%%%%%%%%%%%%%%%%%%%%%%%%%%
%\section*{APPENDIX}


\section*{ACKNOWLEDGMENT}
The authors would like to thank Luyun Qin (SRCB-Ricoh) for his work in data preprocessing.

\bibliographystyle{./bibliography/IEEEtran}
\bibliography{./bibliography/IEEEabrv,./bibliography/IEEERef}

\end{document}